\PassOptionsToPackage{numbers}{natbib}
\documentclass{article}

\usepackage[preprint]{neurips_2026}

\usepackage[utf8]{inputenc}
\usepackage[T1]{fontenc}
\usepackage{microtype}
\usepackage{amsmath}
\usepackage{amssymb}
\usepackage{amsfonts}
\usepackage{nicefrac}
\usepackage{booktabs}
\usepackage{graphicx}
\usepackage{caption}
\usepackage{subcaption}
\usepackage{float}
\usepackage{placeins}
\usepackage{url}
\usepackage{xcolor}
\usepackage[hidelinks]{hyperref}

\graphicspath{{figures/}}

\title{From Token Lists to Graph Motifs: Weisfeiler–Lehman Analysis of Sparse Autoencoder Features}

\author{
  Ruben Fernandez-Boullon\textsuperscript{*} \\
  University of Vigo \\
  \texttt{ruben.fernandez.boullon@uvigo.gal}
  \And
  Pablo Magarinos-Docampo \\
  University of Vigo \\
  \texttt{pablo.magarinos@uvigo.gal}
  \And
  Javier Perez-Robles \\
  University of Vigo \\
}

\begin{document}

\maketitle

\begin{abstract}
Sparse autoencoders (SAEs) have become central to mechanistic interpretability, decomposing transformer activations into monosemantic features \cite{toy-models,scaling-sae,bricken}. Yet existing analyses characterise features almost exclusively through top-activating token lists or decoder weight vectors, leaving the higher-order co-occurrence structure shared across features largely unexamined. We introduce a graph-structured representation in which each SAE feature is modelled as a token co-occurrence graph: nodes are the tokens most frequent near strong activations, and edges connect pairs that co-occur within local context windows. A custom WL-style, frequency-binned graph kernel \cite{wl,shervashidze} then provides a similarity measure over this structural space. Applied as a proof of concept to features from a large SAE trained on GPT-2 Small \cite{gpt2} and probed with a synthetic mixed-domain corpus, our clustering recovers heuristic motif families---punctuation-heavy patterns, language and script clusters, and code-like templates---that are not recovered by clustering on decoder cosine similarity (alphabetic purity 0.516 vs.\ 0.000). A token-histogram baseline achieves higher overall purity (0.854 vs.\ 0.760), so the contribution of the graph view is complementary rather than dominant: it surfaces structural relationships that token-frequency and decoder-weight views alone do not capture. Cluster assignments are stable across graph-construction hyperparameters and random seeds.
\end{abstract}

\section{Introduction}

Large language models (LLMs) are increasingly deployed in high-stakes applications, motivating a growing body of work on mechanistic interpretability that aims to understand their internal computation at the level of neurons, features, and circuits. Sparse autoencoders (SAEs) over transformer activations have been proposed as a way to disentangle polysemantic neurons into more monosemantic features, and have been scaled to full residual streams in GPT-2 Small and to frontier models such as Claude 3 Sonnet~\cite{toy-models,bricken,scaling-sae,scaling-mono}. Despite this progress, most empirical analyses of SAEs describe individual features through lists of top-activating tokens, example prompts, or human-written explanations, leaving open the question of how to systematically organise and compare large dictionaries of features.

In this paper, we explore a complementary perspective in which each SAE feature is represented as a graph derived from its activation patterns over text. Intuitively, we treat the tokens that frequently appear in the neighbourhood of strong activations as nodes, and connect pairs of tokens that co-occur in local contexts when the feature is active. This yields a collection of graphs, one per feature, which can be compared using a graph kernel inspired by the Weisfeiler-Lehman (WL) subtree kernel \cite{wl,shervashidze}. By clustering features in this graph-kernel space, we aim to surface recurring structural motifs that group together features which appear unrelated if inspected only via their top tokens.

We propose graph-structured co-occurrence representations for SAE features and a custom WL-style, frequency-binned kernel tailored to weighted token graphs (Sections~\ref{sec:method-graphs}--\ref{sec:method-kernel}). (ii) We provide a proof-of-concept evaluation on a single SAE (6-RES-JB, layer~6) of GPT-2 Small, using a synthetic mixed-domain corpus designed to elicit varied surface motifs (Section~\ref{sec:setup}). (iii) We benchmark the resulting clusters against decoder-cosine and token-histogram baselines using a heuristic token-type purity metric (Section~\ref{sec:results-rq2}), and report robustness to graph-construction hyperparameters, k-means seeds, feature-selection cutoffs, undirected-vs-directed edges, and a small Python-code corpus. We do not claim mechanistic interpretation of individual features; the conclusions are about whether the graph view yields a structural organisation distinct from existing baselines.

\section{Background}

Sparse autoencoders over transformer activations have been proposed as a way to discover more interpretable feature bases than raw neurons, by learning an overcomplete dictionary with sparse codes that reconstruct the original activations \cite{toy-models,scaling-sae}. Recent work has demonstrated that such SAEs can capture meaningful concepts, from simple syntax and punctuation patterns to more abstract semantic and factual knowledge, and can be scaled across all residual stream layers of GPT-2 Small \cite{scaling-sae,bricken}.

Graph kernels provide a way to compare graphs by implicitly mapping them into a feature space where inner products correspond to measures of structural similarity. The Weisfeiler-Lehman (WL) framework iteratively refines node labels by hashing together local neighbourhood information; the resulting WL subtree kernel of Shervashidze et al.~\cite{shervashidze} extends the original combinatorial test of Weisfeiler and Lehman~\cite{wl} into a family of efficient graph kernels widely used for graph classification. Standard Python libraries such as GraKeL~\cite{grakel} and graphkit-learn~\cite{graphkit} provide off-the-shelf implementations. In our experiments we use a custom implementation that departs from the standard subtree kernel in two ways: it derives initial node labels from log-scaled, binned co-occurrence frequencies rather than raw token identities, and it normalises by the geometric mean of diagonal kernel values (Section~\ref{sec:method-kernel}). For brevity we still abbreviate this kernel as ``WL-style'' or simply ``WL'' throughout, with the explicit caveat that it is an inspired variant rather than the kernel of \cite{shervashidze} verbatim.

\section{Related Work}

The superposition hypothesis~\cite{toy-models} motivates SAEs as a tool for decomposing polysemantic neurons into monosemantic features by learning overcomplete dictionaries with sparse latent codes. Bricken et al.~\cite{bricken} established the foundational methodology on a one-layer transformer, recovering hundreds of interpretable features via dictionary learning on MLP activations. At frontier scale, Templeton et al.~\cite{scaling-mono} showed that SAEs trained on Claude 3 Sonnet yield millions of interpretable features spanning fine-grained syntax, factual knowledge, and abstract concepts. Early SAEs impose $L_1$ regularisation, but the resulting shrinkage bias systematically underestimates feature magnitudes~\cite{scaling-sae}. Gated SAEs~\cite{gated-sae} address this by decoupling feature selection from magnitude estimation via a dedicated sigmoid gate and an independent ReLU branch. TopK SAEs~\cite{scaling-sae} replace continuous regularisation with a hard cardinality constraint retaining exactly $K$ activations per forward pass. JumpReLU SAEs~\cite{jumprelu} introduce a learnable per-feature threshold with a discontinuous activation function, enabling direct $L_0$ optimisation via straight-through estimators and achieving state-of-the-art reconstruction fidelity on frontier models. The SAE we analyse (6-RES-JB) uses a standard ReLU activation; whether graph-structured motifs persist across these newer architectural variants is an open direction.

Li et al.~\cite{geometry-concepts} demonstrated through large-scale geometric analysis that SAE feature spaces exhibit rich multi-scale structure rather than being isotropic. At the finest scale, feature vectors form near-crystalline arrangements in latent subspaces that encode relational analogies akin to Word2Vec arithmetic. At an intermediate scale, co-occurrence topology reveals tightly localised ``lobes'' of features sharing semantic domains---such as Python programming, biomedical vocabulary, or French text---that are physically adjacent in decoder geometry. At the broadest scale, the spectral distribution of the full feature point cloud follows a power law, with maximum variance concentrated in middle transformer layers. This multi-scale geometric organisation directly motivates the graph-based view of this paper: rather than describing features in isolation through top-activating tokens, we build co-occurrence graphs that capture neighbourhood structure across activations and apply graph kernels to surface higher-order motifs spanning multiple features simultaneously.

Mechanistic interpretability has moved toward explicitly graph-structured representations of information flow. Marks et al.~\cite{sparse-circuits} introduced sparse feature circuits---directed acyclic graphs connecting SAE features across layers via causal attribution---and showed that these circuits provide concise, editable explanations of model behaviour on narrow tasks. Building on this, Dunefsky et al.~\cite{transcoders} proposed Cross-Layer Transcoders (CLTs), which replace opaque MLP blocks with linear surrogates projecting residual states to post-MLP outputs as a direct linear combination of learned features, enabling scalable construction of full attribution graphs in which edges are weighted by Jacobian-based causal influence. To our knowledge, no prior work applies graph kernels to compare SAE feature co-occurrence graphs; our approach is complementary to circuit tracing, operating at the level of local activation structure rather than inter-layer causal attribution.

\section{Method}

\subsection{Models and sparse autoencoders}

We focus on GPT-2 Small as our base language model, using the implementation and pretrained weights from HuggingFace Transformers \cite{gpt2}. For sparse autoencoders, we rely on the public open-sae-soup collection \cite{scaling-sae,bricken} trained on the residual stream of GPT-2 Small. Concretely, all experiments in this paper use a single SAE attached to the residual stream at layer~6 before the attention block (6-RES-JB), which exposes $24{,}576$ sparse features with an expansion factor of~32 ($d_\mathrm{model} = 768$, $d_\mathrm{sae} = 24{,}576$). The SAE was originally trained on an OpenWebText-like corpus~\cite{openwebtext}; in this paper we re-use the released weights without retraining and probe activations on a different (synthetic mixed-domain) corpus described in Section~\ref{sec:setup}. We do not run cross-layer comparisons: extending the analysis to other residual layers and SAE architectures is left to future work and discussed as a scope limitation in Section~\ref{sec:limitations}.

\subsection{Constructing graphs from SAE features}\label{sec:method-graphs}

Given a fixed SAE and a corpus of text, we collect activations for a subset of features over many model runs. For each feature $f$, we define a high-activation event at position $t$ whenever its activation $z_{t,f}$ exceeds a feature-specific threshold $\tau_f$, chosen as the $p$-th percentile of its empirical activation distribution. Around each such event, we consider a local window of tokens and accumulate statistics about which tokens tend to appear in the neighbourhood of strong activations.

To convert feature $f$ into a graph $G_f$, we proceed as follows. First, we select the top $K$ tokens that appear most frequently in windows centred around high-activation events for $f$; these tokens form the node set of $G_f$. Each node carries a discrete label derived from the log-scaled co-occurrence frequency of its token within the feature (Section~\ref{sec:method-kernel}), so node identity reflects activation strength rather than token identity. Second, we add an undirected edge between two nodes if their corresponding tokens co-occur at least $C$ times within such windows for feature $f$, with edge weight equal to the co-occurrence count. This yields a weighted graph that captures the local co-activation structure induced by the feature across many contexts. In addition to the primary graph specification, we later test the robustness of our conclusions to reasonable variations in the window size, top-$K$ token cutoff, and co-occurrence threshold used to instantiate these graphs. We note that GPT-2's autoregressive architecture implies a natural temporal ordering of tokens; we therefore also evaluate a directed variant of these graphs (Appendix~\ref{app:directed}) in which edges respect a frequency-based proxy for causal ordering.

\subsection{Comparing features with a WL-style graph kernel}\label{sec:method-kernel}

Once each feature is represented as a labelled weighted graph $G_f$, we compare pairs of features using a custom WL-style graph kernel. We perform $h = 3$ refinement iterations starting from initial node labels derived from log-scaled co-occurrence frequencies ($\log(1 + c)$, where $c$ is the row-sum of co-occurrences for each node), discretised into 64 bins. At each iteration, every node label is updated as a weighted average of its neighbours' labels (weighted by edge co-occurrence counts), and re-binned. The resulting per-graph histogram of node-label counts is treated as an explicit feature map, and the kernel value $K(G_f, G_{f'})$ is the inner product of these histograms. This implementation departs from the canonical WL subtree kernel of Shervashidze et al.~\cite{shervashidze} in three respects: (i) the initial labels encode co-occurrence strength rather than raw token identity; (ii) label refinement aggregates neighbour information through weighted averaging and binning, rather than label hashing; and (iii) the final kernel sums over a single histogram rather than concatenating per-iteration histograms. We therefore refer to it as a ``WL-style frequency-binned graph kernel'' rather than as the standard WL subtree kernel.

The resulting kernel matrix $K \in \mathbb{R}^{N \times N}$ for $N$ features captures shared local neighbourhood structure. We normalise $K$ to $[0, 1]$ by dividing each entry $K_{ij}$ by the geometric mean of the diagonal entries $\sqrt{K_{ii} K_{jj}}$, then symmetrise as $K \leftarrow (K + K^\top)/2$. We embed features into a low-dimensional space by applying kernel PCA with the precomputed kernel to 2 dimensions (\texttt{random\_state=42}), and perform clustering via k-means (\texttt{n\_init=20}, \texttt{random\_state=42}) in this embedding space to obtain groups of features that share similar structural motifs.

Throughout, we compare this WL-style representation to two baselines: (i) clustering in the space of SAE decoder vectors using cosine similarity, and (ii) clustering based on token histograms constructed from each feature's top-$K$ tokens.

\subsection{Token-type labels and cluster purity}\label{sec:method-purity}

To evaluate whether clusters correspond to interpretable token-type families, we use a heuristic labelling scheme and a standard cluster-purity metric, both of which we summarise here and define formally in Appendix~\ref{app:purity}. Each token $t$ is decoded to a string and assigned a coarse label $\ell(t) \in \{\text{symbolic}, \text{alphabetic}, \text{numeric}, \text{mixed}\}$ based on the dominant character class of its decoded form (Python \texttt{string.punctuation}/\texttt{digits} for symbolic, ASCII letters for alphabetic, digits for numeric, and \emph{mixed} for tokens that fall in none of these majority classes). Each feature $f$ then receives the plurality label $\hat\ell(f)$ over its top-$K$ tokens. Given a clustering $\{I_c\}_{c=1}^{K}$, the dominant label $\hat y_c$ of cluster $c$ is the plurality of $\hat\ell$ over its members, the per-cluster purity is the fraction of members whose label matches $\hat y_c$, and the overall purity~$P$ is the size-weighted average:
\begin{equation}
  P = \frac{1}{N} \sum_{c=1}^{K} \bigl|\{i \in I_c : \hat\ell(i) = \hat y_c\}\bigr|.
  \label{eq:purity}
\end{equation}
Category-specific purities (e.g.\ alphabetic, symbolic) average per-cluster purity over clusters whose dominant label equals the target category.

\section{Experimental Setup}\label{sec:setup}

\subsection{Data}

Our main experiments use a synthetic mixed-domain corpus whose released snapshot contains \texttt{n\_tokens\_collected = 78{,}749} token positions (approximately 79k), generated procedurally from a fixed set of templates spanning $13$ surface registers: Python and JavaScript code, Bash commands, mathematical expressions, URLs and file paths, JSON structures, social-media-style posts, natural-language paragraphs, formal e-mail headers, and short passages in Spanish, French, German, and Japanese. The mixture is constructed (with random seed~$42$) to elicit a wide range of surface motifs---punctuation density, alphabetic structure, code syntax, multi-script tokens---so that any motif families surfaced by the WL-style kernel are not artefacts of a single text register. The full templates, domain weights, and exact token budget are listed in the released code (\texttt{main.py}). Tokens are encoded with the GPT-2 byte-pair tokeniser; the corpus is fed through GPT-2 Small~\cite{gpt2} with the 6-RES-JB SAE hooked at layer~6 to record activations for all selected features.

This synthetic-corpus design is a deliberate scope choice: the SAE itself was trained on a real OpenWebText-like corpus, but our analysis only inspects how its features partition across texts we feed in. Using a templated corpus exchanges naturalistic frequency statistics for explicit control over which surface registers are present, at the cost of distributional realism. Generalisation to natural distributions such as OpenWebText~\cite{openwebtext} or to a code corpus such as The Stack~\cite{thestack} therefore remains an empirical question; we report a small Python-code spot check in Appendix~\ref{app:cross-corpus} and discuss this limitation explicitly in Section~\ref{sec:limitations}.

\subsection{Feature selection and graph construction}

From the full set of $24{,}576$ features in the SAE, we select a subset of $N = 2048$ features for analysis, retaining features whose nonzero activation fraction lies between $0.1\%$ and $98\%$. For each selected feature we build a co-occurrence graph as described in Section~\ref{sec:method-graphs}, yielding a collection of $N = 2048$ labelled graphs. Each feature is also assigned a heuristic token-type label as defined in Section~\ref{sec:method-purity}, which is used to compute cluster purity scores. The sensitivity of clustering results to the choice of $\alpha_{\min}$ and $\alpha_{\max}$ is examined in Section~\ref{sec:cutoff-ablation}.

\subsection{Kernel configuration and baselines}

For the WL-style kernel, we use a custom implementation (not GraKeL~\cite{grakel} or graphkit-learn~\cite{graphkit}) with the configuration introduced in Section~\ref{sec:method-kernel}: $h = 3$ refinement iterations, log-scaled initial node labels, 64 label bins per iteration, and geometric-mean normalisation to $[0, 1]$. We compute the full $N \times N$ kernel matrix and symmetrise it before applying kernel PCA and clustering.

For baselines, we construct two alternative similarity matrices: (i) cosine similarity between SAE decoder vectors (normalised to unit $L_2$ norm), and (ii) cosine similarity between normalised token histograms for each feature, constructed from the top-$K$ token counts per feature.

\subsection{Complete parameter specification}

All experimental parameters are listed below for reproducibility. \textbf{SAE:} 6-RES-JB from open-sae-soup~\cite{scaling-sae}, layer 6 residual stream (\texttt{resid\_pre}), $d_\mathrm{model} = 768$, $d_\mathrm{sae} = 24{,}576$, expansion factor 32, ReLU activation. \textbf{Corpus:} synthetic mixed-domain corpus described in Section~\ref{sec:setup}, generated with seed 42. \textbf{Token budget:} $M = 78{,}749$ token positions. \textbf{Feature selection:} percentile threshold $p = 50$, min nonzero $= 0.1\%$, max nonzero $= 98\%$, yielding $N = 2048$ features. \textbf{Graph:} window size $W = 10$, top-$K$ tokens $=30$, co-occurrence threshold $C = 3$, minimum $5$ high-activation events, max $200$ events per feature. \textbf{Kernel:} $h = 3$ WL-style iterations, 64 bins, log-scaled labels, geometric-mean normalisation. \textbf{Clustering:} k-means ($K = 10$ clusters, \texttt{n\_init=20}, \texttt{random\_state=42}). \textbf{Kernel PCA:} 2 components, \texttt{random\_state=42}. \textbf{Motif example selection:} feature closest to cluster centroid in the kernel PCA embedding. \textbf{Seed:} all random states $=42$ unless reported otherwise.

To test whether the clustering structure depends strongly on a single graph-construction choice, we evaluate a small grid of reasonable hyperparameters on the primary SAE layer and main corpus. We compare window sizes $W \in \{5, 10, 15\}$, top-token cutoffs $K \in \{30, 50\}$, and co-occurrence thresholds $C \in \{3, 5\}$, while keeping the remaining parts of the pipeline fixed. We compare the resulting cluster assignments using stability metrics: Adjusted Rand Index (ARI) and Normalised Mutual Information (NMI), measured relative to the default configuration and across pairs of settings.

\section{Results}

\subsection{RQ1: Do graph-based clusters correspond to structural motif families?}

\begin{figure}[htbp]
    \centering
    \begin{minipage}[t]{0.46\textwidth}
        \centering
        \includegraphics[width=\textwidth]{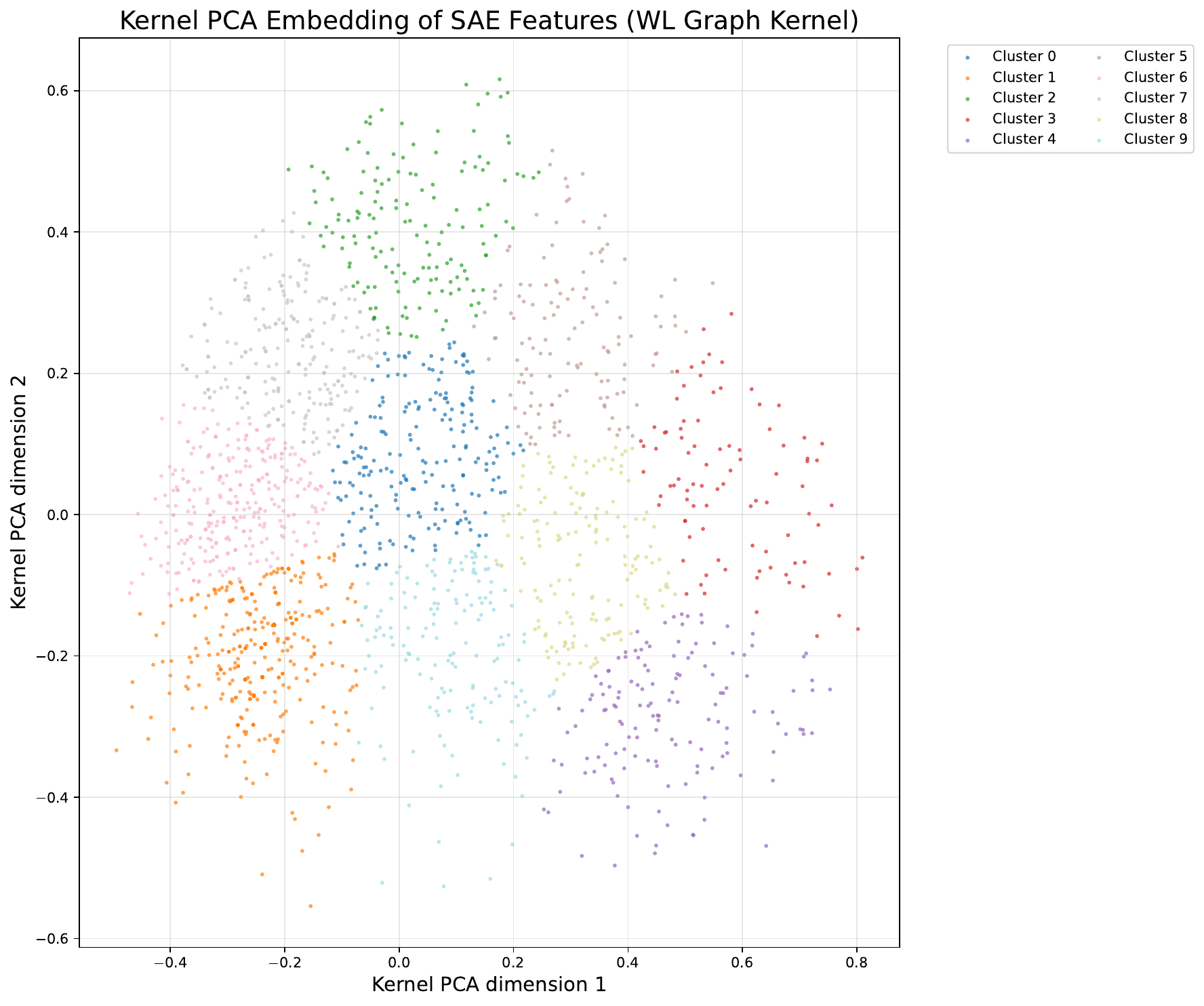}
        \subcaption{Kernel PCA embedding coloured by k-means cluster assignment ($K = 10$).}
        \label{fig:wl-embedding}
    \end{minipage}\hfill
    \begin{minipage}[t]{0.50\textwidth}
        \centering
        \includegraphics[width=\textwidth,trim=0 3439.8bp 0 0,clip]{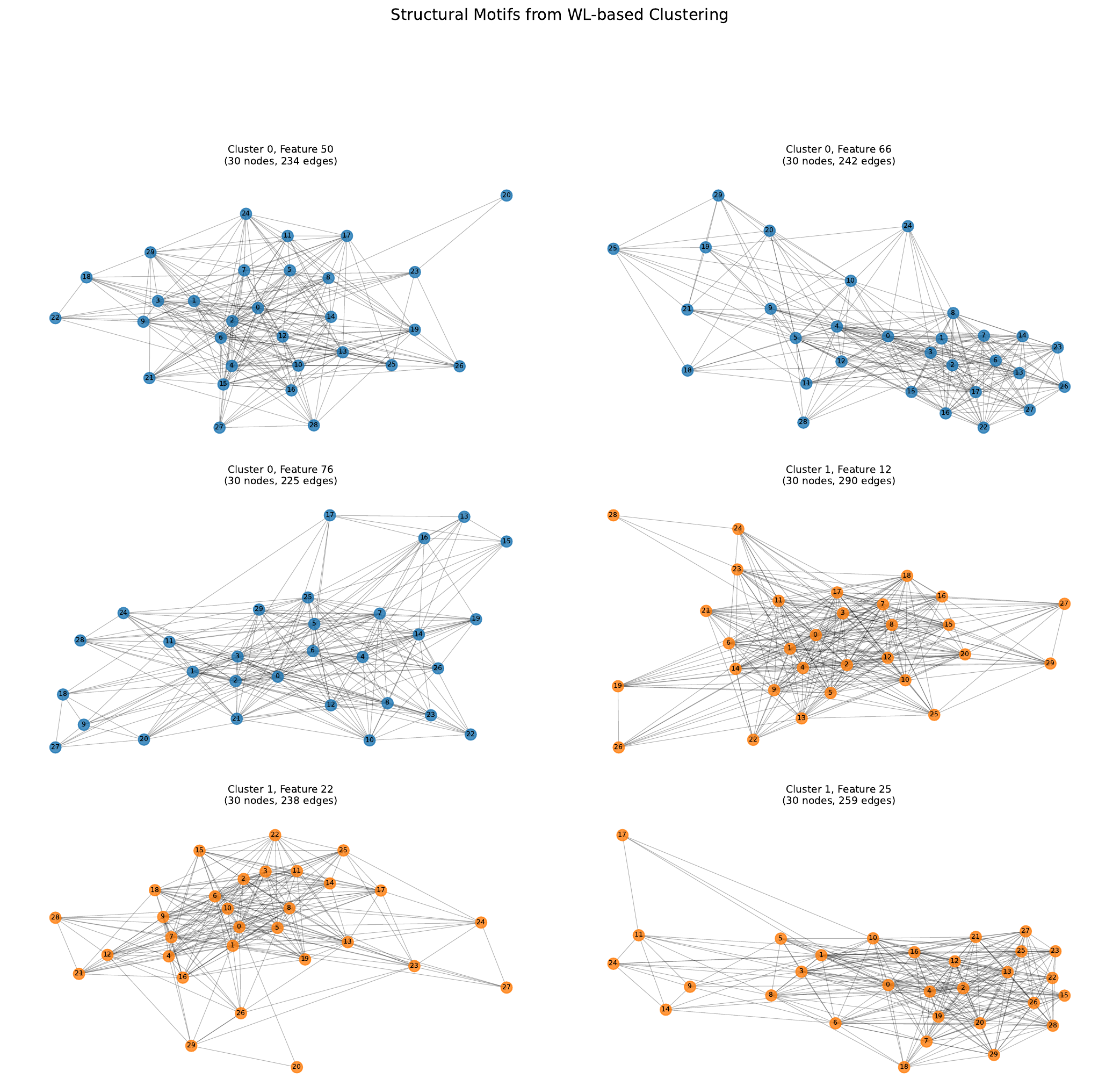}
        \subcaption{Representative co-occurrence graphs from selected clusters.}
        \label{fig:graph-motifs}
    \end{minipage}
    \caption{(a) Kernel PCA embedding of the $N = 2048$ selected SAE features from layer~6 (6-RES-JB), based on the custom WL-style frequency-binned graph kernel ($h = 3$, $W = 10$, $K = 30$, $C = 3$), coloured by k-means cluster assignment ($n_\mathrm{init} = 20$). The $2048$ features are selected from the $24{,}576$ SAE features at activation-percentile threshold $p = 50$ (Section~\ref{sec:setup}). (b) Representative co-occurrence graphs drawn from the WL-style clusters, illustrating qualitatively distinct surface motifs (punctuation-heavy patterns, code-like structures, language- and script-specific token communities). A larger gallery is provided in the released code.}
\end{figure}

To obtain a global view of the feature space, we embed the WL-style kernel matrix into two dimensions using kernel PCA and apply k-means clustering with $K = 10$ clusters. Figure~\ref{fig:wl-embedding} shows the resulting 2D embedding, coloured by cluster assignment. Under the heuristic token-type labels of Section~\ref{sec:method-purity}, features form visually coherent regions that we interpret as broadly distinct motif families: punctuation-heavy patterns, code-like structures with operator and bracket tokens, and a smaller set of multi-script and natural-language token clusters.

To make these motifs more concrete, we inspect several clusters in detail. For each cluster, we select a few prototype features closest to the cluster centroid in the embedding space, and visualise their graphs along with their top-activating tokens. Figure~\ref{fig:graph-motifs} shows representative graphs, illustrating tightly connected punctuation subgraphs, mixed alphanumeric patterns reminiscent of identifiers, and loosely connected token communities tied to specific languages or domains. These visualisations are illustrative rather than exhaustive: a quantitative comparison against baselines is given in Section~\ref{sec:results-rq2}, and a small-scale verification on a Python code corpus in Appendix~\ref{app:cross-corpus}.

\FloatBarrier

\subsection{RQ2: Does the graph view capture structure beyond token and decoder baselines?}\label{sec:results-rq2}

We next compare the WL-style clustering to two simpler baselines: clustering in the space of SAE decoder vectors, and clustering based on token histograms. To quantify the extent to which clusters align with simple heuristics, we automatically assign each feature a coarse label based on the composition of its top-activating tokens, distinguishing predominantly symbolic, alphabetic, numeric, and mixed patterns. We then compute cluster purity (Section~\ref{sec:method-purity}, Eq.~\ref{eq:purity}) with respect to these heuristic labels under each similarity measure.

Table~\ref{tab:cluster-purity} reports the average cluster purity for WL-style, decoder-based, and token-histogram-based similarity.

In this evaluation, WL captures structural regularities not reflected in decoder vectors alone: it achieves an alphabetic purity of $0.516$ versus $0.000$ for decoder cosine, indicating that under a $K=10$ k-means partition the decoder-cosine clustering does not isolate any cluster whose plurality token-type label is alphabetic, whereas the WL-style clustering does. WL also achieves a slightly higher overall purity than decoder cosine ($0.760$ vs.\ $0.754$) and a higher symbolic purity ($0.781$ vs.\ $0.754$). The token histogram baseline, however, achieves the highest overall purity ($0.854$) by directly leveraging token-frequency information. We therefore read these numbers as showing that the token-histogram baseline is a strong competitor for overall feature organisation, and that WL's marginal advantage lies in capturing alphabetic structural patterns that survive after factoring out marginal token frequencies, rather than in dominating broad clustering quality. Differences this small (e.g., $0.760$ vs.\ $0.754$) should be read with caution: we report multi-seed variability below but do not compute bootstrap confidence intervals over the underlying feature sample.

\begin{table}[H]
    \centering
    \begin{tabular}{lccc}
        \toprule
        Method & Symbolic & Alphabetic & Overall \\
        \midrule
        WL-style kernel  & 0.781 & 0.516 & 0.760 \\
        Decoder cosine   & 0.754 & 0.000 & 0.754 \\
        Token histogram  & 0.894 & 0.749 & 0.854 \\
        \bottomrule
    \end{tabular}
    \caption{Cluster purity (Eq.~\ref{eq:purity}) with respect to heuristic token-type labels under our WL-style frequency-binned graph kernel, decoder-cosine similarity, and top-$K$ token-histogram cosine similarity. Higher is more aligned with the heuristic labels; differences below $\sim$$0.01$ are within multi-seed noise (Table~\ref{tab:ablation}).}
    \label{tab:cluster-purity}
\end{table}

Per-cluster sizes (range $91$--$484$) and per-cluster purity values are listed in Appendix~\ref{app:cluster-summary} (Table~\ref{tab:cluster-summary}). The dominant type is symbolic (punctuation-heavy motifs) for nine of ten clusters; cluster 8 is the sole alphabetic-dominant cluster, with purity $51.6\%$.

\subsection{Robustness to graph construction hyperparameters}

We next examine how sensitive the clustering is to the graph-construction hyperparameters introduced in Section~\ref{sec:method-graphs}. Across all $3 \times 2 \times 2 = 12$ grid points with $W \in \{5, 10, 15\}$, $K \in \{30, 50\}$, and $C \in \{3, 5\}$, overall purity ranges from $0.655$ to $0.812$ (mean $0.748 \pm 0.051$, default $W=10$, $K=30$, $C=3$). Qualitatively, the main motif families are preserved: symbol clusters stay tightly grouped, language and script communities still form distinct regions, and code-like motifs continue to appear as separate families. We read this as evidence that the graph-structured view is not an artefact of a single window size or sparsification threshold, although the exact cluster boundaries do shift modestly under more aggressive or more permissive settings.

\subsection{Ablation study and multi-seed stability}

We assess sensitivity to k-means initialisation by re-running clustering with $10$ random seeds ($\{42, 0, 7, 13, 21, 33, 47, 64, 99, 123\}$) on the WL-style embedding. Mean overall purity across seeds is $0.7573 \pm 0.0019$, with $\sigma < 0.01$ for every metric in Table~\ref{tab:ablation}, indicating stability with respect to initialisation. We further ablate across seven similarity measures: the default WL-style kernel, the same kernel with edge weights zeroed, the same kernel with node labels randomly shuffled, decoder cosine, token histogram, co-occurrence-matrix cosine, and Jaccard similarity on top-token sets.

\begin{table}[H]
    \centering
    \begin{tabular}{lrrr}
        \toprule
        Method & Overall & Alphabetic & ARI vs.\ default \\
        \midrule
        WL-style (default)        & $0.757$ & $0.500$ & -- \\
        WL-style, edges removed   & $0.754$ & $0.000$ & $0.088$ \\
        WL-style, labels shuffled & $0.768$ & $0.584$ & $0.618$ \\
        Decoder cosine            & $0.754$ & $0.000$ & $0.002$ \\
        Token histogram           & $0.853$ & $0.755$ & $0.208$ \\
        Co-occurrence cosine      & $0.769$ & $0.729$ & $0.030$ \\
        Jaccard top-token         & $0.839$ & $0.658$ & $0.190$ \\
        \bottomrule
    \end{tabular}
    \caption{Ablation with multi-seed k-means (10 seeds; $\sigma \leq 0.01$ for every column, omitted for space). Zeroing the edges of the WL-style kernel drops alphabetic purity to $0$ while leaving overall purity essentially unchanged: edges (not just node-label histograms) are what give the WL view its alphabetic-dominant cluster. Shuffling node labels keeps purity above the default and yields ARI $0.618$ with the default, indicating that the kernel relies mostly on edge structure and marginal label frequencies, not specific label identities. Token-histogram and co-occurrence-matrix baselines also achieve high alphabetic purity, so much of the broader signal is recoverable from token-frequency or co-occurrence marginals.}
    \label{tab:ablation}
\end{table}

The headline observation is the gap on alphabetic purity between the WL-style kernel ($0.500$) and decoder cosine ($0.000$); we report it as an empirical observation under the heuristic metric, not as a statistically tested claim. The reported point estimates are stable to k-means initialisation but remain artefacts of a single feature sample, a single SAE layer, and a single probing corpus.

\subsection{Sensitivity to feature selection cutoffs}
\label{sec:cutoff-ablation}

We additionally probe whether the clustering depends on the specific activation-fraction cutoffs $\alpha_{\min}, \alpha_{\max}$ used to select the $N=2048$ features (Section~\ref{sec:method-graphs}). Sweeping five configurations spanning $[\alpha_{\min}, \alpha_{\max}] \in \{[0.001,0.98], [0.005,0.95], [0.010,0.90], [0.001,0.99], [0.0005,0.98]\}$ keeps mean purity in the range $0.7547$--$0.7573$ with $\sigma \leq 0.002$ across $10$ k-means seeds, all within multi-seed noise of the baseline (full table in Appendix~\ref{app:cutoff-ablation}). The $0.1\%$--$98\%$ thresholds therefore do not appear to introduce a material selection bias.

\subsection{Directed graphs and cross-corpus spot checks}\label{sec:directed-and-cross}

Two further checks, deferred to the appendix for space, probe robustness along orthogonal axes. (i)~\textbf{Directed graphs} (Appendix~\ref{app:directed}): replacing each undirected co-occurrence graph with a lower-triangular approximation that mimics autoregressive ordering preserves overall purity ($0.754$ vs.\ undirected $0.757$) but eliminates the alphabetic-dominant cluster (ARI $0.123$ vs.\ undirected). The triangular approximation discards roughly half of the co-occurrence signal, and the sparser alphabetic structure appears to fall below the threshold needed for cluster formation; a fully causal treatment requires positional data not retained by the cached graphs used here. (ii)~\textbf{Python-code spot check} (Appendix~\ref{app:cross-corpus}): re-running the kernel and clustering on a generated Python corpus of order $10^4$--$10^5$ tokens yields $99.1\%$ feature overlap with the main run and ARI/NMI of $0.944$/$0.928$ against the main-corpus clustering, so the cluster structure is largely stable when activations are recomputed on a focused code corpus. Within the main clustering, clusters~$6$ and~$7$ carry the highest density of Python keywords, operators, and brackets ($33.1\%$ and $32.5\%$ of top tokens), consistent with code-like motifs being concentrated in specific clusters. We treat both checks as suggestive rather than definitive; an evaluation on a naturalistic code corpus such as The Stack~\cite{thestack} remains future work.

\section{Discussion}\label{sec:discussion}

WL-style kernels aggregate local neighbourhood information: what matters is not only which tokens appear near a feature's activations, but how those tokens are connected within the same windows. In our setting, alphabetic patterns appear to produce topological signatures---densely connected token communities in predictable sequential arrangements---that the graph view reflects but decoder cosine does not. Decoder cosine captures what a feature ``points toward'' in residual-stream geometry, but is blind to the co-occurrence structure of activation contexts; this is consistent with its collapse to $0.000$ alphabetic purity in our $K=10$ partition. We read this as a possible (not causally established) distinction between decoder geometry and activation-context topology, related to but weaker than the multi-scale geometric organisation reported by Li et al.~\cite{geometry-concepts}.

The higher overall purity of the token histogram ($0.854$ vs.\ $0.760$ for WL) is consistent with a simple property of layer-6 features: marginal token distributions already capture most of the information needed for the broad heuristic categorisation we use. The graph view appears to add information specifically where the \emph{relationships} among co-occurring tokens matter, most visibly for alphabetic features (WL: $0.516$ vs.\ histogram: $0.749$, decoder: $0.000$; Table~\ref{tab:cluster-purity}). The dominance of symbolic clusters ($9$ of $10$) with a single alphabetic cluster is consistent with layer~6 carrying substantial positional and syntactic information, and suggests that graph-based analysis is most informative as a complement to token-frequency methods rather than as a replacement.

\paragraph{Broader impact.}
This work is foundational interpretability research that analyses existing, publicly available language models and sparse autoencoders. Positive societal impacts include improved transparency and auditability of neural language models: by systematically organising the internal representations of transformer models, this line of research contributes tools that could help practitioners identify potentially harmful or biased feature clusters, facilitate safety auditing of AI systems, and improve human understanding of model behaviour in high-stakes applications. We identify no direct negative societal impact: the paper does not introduce new generative capabilities, deployment-ready systems, or high-risk model releases. As with any interpretability methodology, a possible indirect risk is that a deeper understanding of model internals could in principle be exploited to reverse-engineer or circumvent safety measures in future systems; we consider this risk low for the present contribution, given its analytical and non-generative character.

\section{Limitations}
\label{sec:limitations}

\paragraph{Scalability.}
The WL-style kernel computation requires forming the full $N \times N$ pairwise kernel matrix, which entails $O(N^2 \cdot h)$ work in the number of selected features $N$ and refinement iterations $h$. For the current setting of $N = 2048$ and $h = 3$ this is tractable on a single CPU. However, scaling to the full $24{,}576$-feature dictionary or to larger modern SAEs with up to $N = 131{,}072$ features~\cite{scaling-sae} would render the dense kernel matrix infeasible in both time and memory. Practical extensions would require approximate kernel methods (e.g., Nystr\"{o}m approximations, random feature maps for graph kernels) or hierarchical clustering strategies that avoid materialising the full matrix.

\paragraph{Custom kernel rather than standard WL.}
The kernel used here departs from the standard WL subtree kernel of Shervashidze et al.~\cite{shervashidze} in three ways (Section~\ref{sec:method-kernel}): initial node labels encode log-binned co-occurrence counts rather than token identities, refinement uses weighted neighbour averaging rather than label hashing, and the final feature map sums over a single iteration's histogram rather than concatenating across iterations. We have not benchmarked the standard WL subtree kernel (e.g., as implemented in GraKeL~\cite{grakel}) against ours; it is therefore an open question whether the alphabetic-cluster gap over decoder cosine survives under a standard WL implementation, or is in part an artefact of our choice of label and refinement schemes. The label-shuffle ablation (Table~\ref{tab:ablation}) suggests that purity is largely insensitive to the specific log-binned identities, mitigating but not eliminating this concern.

\paragraph{Causal graph structure.}
Our default graph construction treats co-occurrence edges as undirected, ignoring GPT-2's autoregressive causal structure. We partially address this in Appendix~\ref{app:directed} with a directed graph variant. Our released code (\texttt{src/graph\_construction.py}) also exposes a \texttt{directed=True} mode that records true positional ordering at graph-construction time; we left the experimental evaluation of that mode to future work, since it requires recomputing the full set of feature graphs.

\paragraph{Scope of evaluation.}
This study focuses on a single base model (GPT-2 Small) and a single SAE architecture (6-RES-JB) at a single residual layer; stronger claims about layerwise or model-family generalisation require broader sweeps. The probing corpus is a synthetic mixed-domain mixture (Section~\ref{sec:setup}) chosen to expose surface-form variation, not a sample from any particular natural distribution; the SAE itself was trained on an OpenWebText-like corpus and we do not retrain or reanalyse it on naturalistic text in this paper. Our evaluation relies on heuristic token-type labels that capture broad surface structure but do not constitute causal or mechanistic proofs about what each feature actually represents; probe-based or human validation would be needed to confirm that motif families correspond to meaningful interpretability categories. The cross-corpus spot check (Appendix~\ref{app:cross-corpus}) uses generated Python templates rather than a naturalistic code corpus; evaluation on The Stack~\cite{thestack} at scale remains an important direction. Finally, small differences in purity between methods (e.g., $0.760$ vs.\ $0.754$) should be interpreted cautiously without bootstrap confidence intervals over the underlying feature sample.

\section{Conclusion}

We introduced a graph-structured perspective on sparse autoencoder features in transformer language models, in which each feature is represented as a weighted graph capturing co-occurrence relationships among tokens in the neighbourhood of its activations. By comparing these graphs with a custom WL-style frequency-binned kernel and clustering the resulting similarity matrix, we surfaced families of features corresponding to heuristic motif categories that are not separated by clustering on decoder vectors alone. Our headline empirical observation is that the graph view recovers an alphabetic-dominant cluster (purity $0.516$) that decoder cosine does not ($0.000$), while a token-histogram baseline outperforms the graph view on overall purity ($0.854$ vs.\ $0.760$). We therefore present graph-based representations as a complementary lens rather than a replacement for token-frequency analyses, and limit our claims to surface-level structural motifs. Future work should validate motif families through human evaluation, scale the analysis to multiple SAEs, layers, and natural corpora, and compute statistical uncertainty (e.g., bootstrap confidence intervals) for purity and similarity metrics. Code is publicly available; the repository URL is anonymised for double-blind review and will be provided in the camera-ready version. Full reproduction steps are given in Appendix~\ref{app:reproducibility}.

\begin{ack}
\end{ack}

\appendix

\section{Reproducibility}\label{app:reproducibility}

All code required to reproduce the experiments will be made publicly available. The entire experimental pipeline can be executed sequentially, covering the following general stages:Model and SAE Initialization: Load the standard GPT-2 Small model alongside the targeted layer-6 Sparse Autoencoder (specifically, the 6-RES-JB SAE from standard open-source repositories).Corpus Generation and Activation Collection: Construct the synthetic mixed-domain corpus using a fixed random seed (42). Process this corpus through the model to collect activations for all $24{,}576$ SAE features across the target token positions (approximately $78{,}749$ tokens, using a standard batch size such as $512$).Feature Selection: Filter the features based on their activation frequency, retaining those within a specific nonzero activation fraction (e.g., $[0.1\%, 98\%]$), and select the top $N = 2048$ features based on activity using a 50th-percentile threshold.Graph Construction: For each selected feature, build a weighted co-occurrence graph using a window size of $W = 10$, considering the top $K = 30$ tokens, and applying a co-occurrence threshold of $C = 3$.Kernel Computation: Compute the $N \times N$ kernel matrix using a Weisfeiler-Lehman-style approach with $h = 3$ iterations, log-scaled labels, $64$ bins per iteration, and geometric-mean normalization.Clustering: Reduce dimensionality using Kernel PCA (2 components) and apply k-means clustering ($K = 10$ clusters, 20 initializations) using fixed random seeds to ensure consistency.Evaluation and Ablations: Compute standard evaluation metrics, including heuristic token-type labels, cluster purity, and ARI/NMI. The pipeline also includes steps to systematically reproduce the robustness grid, ablation studies (such as multi-seed checks, similarity measures, and cutoff variations), and cross-corpus analyses discussed in the main text and appendices.DependenciesThe experimental framework requires Python 3.10 or higher and relies on standard machine learning, scientific computing, and graph processing libraries (including PyTorch, Transformers, Accelerate, NumPy, SciPy, scikit-learn, NetworkX, GraKeL, and pandas). All random seeds are set to $42$ by default to ensure exact reproducibility across runs.

\section{Heuristic token-type labels and cluster purity}\label{app:purity}

This appendix gives the formal definitions summarised in Section~\ref{sec:method-purity}.

\paragraph{Character-type partition.}
We partition the Unicode code points into three (overlapping) sets, derived from Python's \texttt{string} module:
\begin{align*}
  \mathcal{S} &= \texttt{string.punctuation} \cup \texttt{string.digits} \cup \{\texttt{' '},\, \texttt{'\textbackslash{}n'},\, \texttt{'\textbackslash{}t'}\}, \\
  \mathcal{A} &= \texttt{string.ascii\_letters}, \\
  \mathcal{N} &= \texttt{string.digits}.
\end{align*}
Note that $\mathcal{N} \subset \mathcal{S}$; digits are counted in both symbolic and numeric tallies.

\paragraph{Token label $\ell(t)$.}
Given a token $t$ decoded to a string $s_t$, define character-type counts
\[
  c_{\mathcal{S}}(t) = \sum_{x \in s_t} \mathbf{1}[x \in \mathcal{S}], \quad
  c_{\mathcal{A}}(t) = \sum_{x \in s_t} \mathbf{1}[x \in \mathcal{A}], \quad
  c_{\mathcal{N}}(t) = \sum_{x \in s_t} \mathbf{1}[x \in \mathcal{N}],
\]
and let $\mathrm{total}(t) = c_{\mathcal{S}}(t) + c_{\mathcal{A}}(t) + c_{\mathcal{N}}(t)$. The character-fraction scores are $\pi_{\mathcal{S}}(t) = c_{\mathcal{S}}(t)/\mathrm{total}(t)$, and analogously for $\pi_\mathcal{A}$ and $\pi_\mathcal{N}$. The token label is assigned by the priority rule
\[
  \ell(t) =
  \begin{cases}
    \text{symbolic}   & \text{if } \pi_{\mathcal{S}}(t) > 0.5, \\
    \text{alphabetic} & \text{else if } \pi_{\mathcal{A}}(t) > 0.5, \\
    \text{numeric}    & \text{else if } \pi_{\mathcal{N}}(t) > 0.3, \\
    \text{mixed}      & \text{otherwise.}
  \end{cases}
\]

\paragraph{Feature label $\hat{\ell}(f)$ and cluster purity.}
For a feature $f$ with top-$K$ tokens $T_f = \{t_1, \ldots, t_K\}$, the feature label is the plurality vote $\hat{\ell}(f) = \arg\max_y \bigl|\{t \in T_f : \ell(t) = y\}\bigr|$, with $y \in \{\text{symbolic}, \text{alphabetic}, \text{numeric}, \text{mixed}\}$. Given a clustering $\{I_c\}_{c=1}^{K}$ of the $N$ features, the dominant label of cluster $c$ is $\hat y_c = \arg\max_y \bigl|\{i \in I_c : \hat\ell(i) = y\}\bigr|$, and per-cluster purity is $\mathrm{purity}(c) = |\{i \in I_c : \hat\ell(i) = \hat y_c\}|/|I_c|$. The overall purity reported in Eq.~\ref{eq:purity} is the size-weighted mean. Category-specific purity for label $y$ averages $\mathrm{purity}(c)$ over $\mathcal{C}_y = \{c : \hat y_c = y\}$:
\[
  P_y = \frac{1}{|\mathcal{C}_y|} \sum_{c \in \mathcal{C}_y} \mathrm{purity}(c).
\]

\section{Per-cluster summary}\label{app:cluster-summary}

Table~\ref{tab:cluster-summary} lists the ten WL-style clusters with their sizes, dominant heuristic token type, and per-cluster purity (corresponding to the partition described in Section~\ref{sec:results-rq2}). Cluster sizes range from $91$ to $484$ features. The dominant type is symbolic for nine of ten clusters; cluster~$8$ is the only alphabetic-dominant cluster, with purity $51.6\%$.

\begin{table}[H]
    \centering
    \begin{tabular}{rcrr}
        \toprule
        Cluster & Size & Dominant & Purity \\
        \midrule
        0 & 207 & symbolic & 63.8\% \\
        1 & 484 & symbolic & 87.4\% \\
        2 & 154 & symbolic & 82.5\% \\
        3 & 91 & symbolic & 79.1\% \\
        4 & 143 & symbolic & 56.6\% \\
        5 & 134 & symbolic & 71.6\% \\
        6 & 289 & symbolic & 92.7\% \\
        7 & 226 & symbolic & 86.7\% \\
        \textbf{8} & \textbf{161} & \textbf{alphabetic} & \textbf{51.6\%} \\
        9 & 159 & symbolic & 49.1\% \\
        \bottomrule
    \end{tabular}
    \caption{Summary of the ten WL-style clusters. Bold row is the only alphabetic-dominant cluster; all others are symbolic (punctuation-heavy).}
    \label{tab:cluster-summary}
\end{table}

\section{Sensitivity to feature-selection cutoffs}\label{app:cutoff-ablation}

Table~\ref{tab:cutoff-ablation} reports the full sweep over activation-fraction cutoffs summarised in Section~\ref{sec:cutoff-ablation}. For each configuration we identify which of the $24{,}576$ SAE features would be selected, intersect that set with our current $2048$ features (for which the WL-style embedding is precomputed), and re-run k-means ($K=10$, $10$ seeds) on the resulting subset of the embedding.

\begin{table}[H]
    \centering
    \begin{tabular}{llrcc}
        \toprule
        Config & Cutoff range $[\alpha_{\min}, \alpha_{\max}]$ & $N$ (intersect) & Mean purity & Std \\
        \midrule
        A (baseline)    & $[0.001,\; 0.98]$ & 2048 & 0.7573 & 0.0019 \\
        B (tighter)     & $[0.005,\; 0.95]$ & 2023 & 0.7563 & 0.0020 \\
        C (tightest)    & $[0.010,\; 0.90]$ & 2013 & 0.7547 & 0.0015 \\
        D (wider upper) & $[0.001,\; 0.99]$ & 2048 & 0.7573 & 0.0019 \\
        E (wider lower) & $[0.0005, 0.98]$  & 2048 & 0.7573 & 0.0019 \\
        \bottomrule
    \end{tabular}
    \caption{Sensitivity of k-means purity to feature-selection cutoffs. $N$ (intersect) is the number of features common to each configuration and the current 2048-feature set. Mean purity and standard deviation are over 10 random seeds.}
    \label{tab:cutoff-ablation}
\end{table}

\section{Directed graph variant}\label{app:directed}

GPT-2 is autoregressive: token $i$ can attend only to positions $j \leq i$. Our default co-occurrence graphs ignore this causal ordering by using undirected edges. To probe whether respecting temporal ordering changes the clustering, we construct a directed variant. The released code (\texttt{src/graph\_construction.py}, \texttt{directed=True}) supports recording true positional ordering at graph-construction time, but rebuilding the full set of feature graphs would invalidate the cached kernel matrix; we therefore use a faster post-hoc approximation here and leave the full re-run to future work.

For each symmetric adjacency matrix we retain only its strict lower triangle ($i > j$). The top-tokens list is ordered by frequency (most frequent at index $0$), so this approximation directs edges from less-frequent to more-frequent tokens, reflecting the intuition that a rarer contextual token tends to precede a frequent trigger token in the autoregressive context window. We then apply a directed variant of the WL-style kernel that aggregates in-neighbour and out-neighbour labels separately for each node. The full implementation is in \texttt{experiments/directed\_graph\_experiment.py}.

\begin{table}[H]
    \centering
    \begin{tabular}{lccc}
        \toprule
        Method & Overall & Alphabetic & ARI vs.\ undirected \\
        \midrule
        Undirected WL-style (default)        & $0.757 \pm 0.002$ & $0.500$ & -- \\
        Directed WL-style (lower-triangular) & $0.754 \pm 0.000$ & n/a    & $0.123$ \\
        \bottomrule
    \end{tabular}
    \caption{Directed versus undirected variants of the WL-style kernel ($10$ seeds, $K=10$ clusters). Alphabetic purity is ``n/a'' because no alphabetic-dominant cluster forms under the directed variant.}
    \label{tab:directed}
\end{table}

The directed variant achieves overall purity $0.754$, comparable to the undirected baseline ($0.757$). However, it fails to produce any alphabetic-dominant cluster across all $10$ seeds: every cluster is dominated by symbolic features. The low ARI ($0.123$) indicates that the two variants partition the feature space quite differently despite similar overall purity: the lower-triangular approximation discards roughly half of the co-occurrence signal. A more complete treatment would require raw positional data during graph construction so that true temporal direction can be assigned to each edge.

\section{Cross-corpus spot check on Python code}\label{app:cross-corpus}

Although the main mixed-domain corpus already contains Python and JavaScript snippets, the paper's claim of ``code-like template motifs'' is worth probing on a focused code corpus. We run two complementary spot checks.

\textbf{Token-type analysis (Analysis A).} We decode each cluster's top tokens with the GPT-2 tokeniser and classify each token as: Python keyword (\texttt{def}, \texttt{class}, \texttt{for}, etc.), operator (\texttt{=}, \texttt{+}, \texttt{*}, etc.), bracket/code-structure character, number, alphabetic, punctuation, or other. The \emph{code-token ratio} per cluster is (Python keywords $+$ operators $+$ brackets) divided by total tokens across all features in the cluster (Table~\ref{tab:cross-corpus-token}).

\begin{table}[H]
    \centering
    \begin{tabular}{rrrrrr}
        \toprule
        Cluster & Size & Code ratio & Keywords & Operators & Brackets \\
        \midrule
        6 & 289 & 0.331 & 503 & 946 & 1417 \\
        7 & 226 & 0.325 & 428 & 755 & 1020 \\
        1 & 484 & 0.324 & 695 & 1445 & 2557 \\
        2 & 154 & 0.317 & 300 & 522 & 643 \\
        3 &  91 & 0.310 & 171 & 296 & 380 \\
        5 & 134 & 0.310 & 291 & 425 & 529 \\
        0 & 207 & 0.309 & 470 & 658 & 790 \\
        8 & 161 & 0.304 & 464 & 483 & 523 \\
        9 & 159 & 0.297 & 433 & 456 & 527 \\
        4 & 143 & 0.294 & 369 & 425 & 469 \\
        \bottomrule
    \end{tabular}
    \caption{Code-token ratio per WL-style cluster (Analysis A). Clusters $6$ and $7$ carry the highest concentration of Python keywords, operators, and brackets ($33.1\%$ and $32.5\%$). Cross-cluster differences are modest, so the result should be read as suggestive rather than conclusive.}
    \label{tab:cross-corpus-token}
\end{table}

\textbf{Small-scale code corpus run (Analysis B).} We generate a Python code corpus of $600$ synthesised snippets drawn from $33$ diverse templates (function definitions, class definitions, loops, list comprehensions, decorators, etc.); the resulting tokenised corpus is on the order of $10^4$--$10^5$ tokens. We run this corpus through GPT-2 Small with the SAE hooked at layer~$6$, apply the same feature-selection criteria as in the main experiments, and intersect the resulting active features with the original $2048$. We subset the WL-style kernel matrix to this intersection, re-run kernel PCA $+$ k-means ($K=10$), and compare the cluster assignments against the main-corpus clustering via ARI and NMI (Table~\ref{tab:cross-corpus-eval}).

\begin{table}[H]
    \centering
    \begin{tabular}{lr}
        \toprule
        Metric & Value \\
        \midrule
        Code-corpus snippets & $600$ \\
        SAE features active on code & $23{,}591$ / $24{,}576$ \\
        Intersection with original $2048$ & $2{,}030$ ($99.1\%$) \\
        ARI vs.\ main-corpus clustering & $\mathbf{0.944}$ \\
        NMI vs.\ main-corpus clustering & $\mathbf{0.928}$ \\
        Code-corpus cluster purity & $0.761$ \\
        \bottomrule
    \end{tabular}
    \caption{Cross-corpus spot check (Analysis B). High ARI ($0.944$) and NMI ($0.928$) against the clustering obtained on the main mixed-domain corpus indicate that the cluster structure is largely stable when activations are recomputed on a focused Python-code corpus. This is a small-scale check on generated Python templates; an evaluation on a naturalistic code corpus such as The Stack~\cite{thestack} remains future work.}
    \label{tab:cross-corpus-eval}
\end{table}

\end{document}